\pdfoutput=1
\documentclass[12pt]{article}
\usepackage{amsmath,amsfonts,amsthm,bm} 
\usepackage{graphicx,psfrag,epsf}
\usepackage{enumerate}
\usepackage{natbib}
\usepackage{hyperref}

\usepackage{url}
\usepackage{chngcntr}
\usepackage{dsfont }
\usepackage[dvipsnames]{xcolor}
\usepackage[utf8]{inputenc}
\usepackage{relsize}
\usepackage{amssymb }
\usepackage{enumitem}
\usepackage[english]{babel}	
\usepackage{upgreek }
\usepackage{multirow}
\usepackage[toc,page]{appendix}
\usepackage{boldline}
\usepackage{caption, subcaption}
\usepackage{booktabs}

\usepackage{xcolor}
\usepackage[linesnumbered,ruled,vlined]{algorithm2e}

\SetCommentSty{mycommfont}

\SetKwInput{KwInput}{Input}                
\SetKwInput{KwOutput}{Output}              

\newtheorem{theorem}{Theorem}

\newcommand{\tf}{\textbf}

\newcommand{\RN}[1]{%
  \textup{\uppercase\expandafter{\romannumeral#1}}%
}

\newcommand{\tet}{\textit}
\newcommand{\tc}{\textsc}

\DeclareMathOperator*{\Min}{argmin}

\usepackage{authblk}

  \title{Elastic Coupled Co-clustering for Single-Cell Genomic Data}
  \author[1]{Pengcheng Zeng}
  \author[1]{Zhixiang Lin \footnote{Corresponding author: \href{zhixianglin@sta.cuhk.edu.hk}{zhixianglin@sta.cuhk.edu.hk}}}

\affil[1]{Department of Statistics, The Chinese University of Hong Kong, HK}
\date{\today}

\begin{document}
  \maketitle
  	
\bigskip
\begin{abstract}
The recent advances in single-cell technologies have enabled us to profile genomic features at unprecedented resolution and datasets from multiple domains are available, including datasets that profile different types of genomic features and datasets that profile the same type of genomic features across different species. These datasets typically have different powers in identifying the unknown cell types through clustering, and data integration can potentially lead to a better performance of clustering algorithms. In this work, we formulate the problem in an unsupervised transfer learning framework, which utilizes knowledge learned from auxiliary dataset to improve the clustering performance of target dataset. The degree of shared information among the target and auxiliary datasets can vary, and their distributions can also be different. To address these challenges, we propose an elastic coupled co-clustering based transfer learning algorithm, by elastically propagating clustering knowledge obtained from the auxiliary dataset to the target dataset. Implementation on single-cell genomic datasets shows that our algorithm greatly improves clustering performance over the traditional learning algorithms. The source code and data sets are available at https://github.com/cuhklinlab/elasticC3
\end{abstract}

\noindent%

\section{Introduction}
Clustering aims at grouping a set of objects such that objects in the same cluster are more similar to each other compared to those in other clusters. It has wide applications in many areas, including genomics, where single-cell sequencing technologies have recently been developed. For the analysis of single-cell genomic data, most clustering methods are focused on one data type: SIMLR \citep{Wang2017}, SC3 \citep{Kise2017}, DIMM-SC \citep{Sun2017}, SAFE-clustering \citep{Yang2018} and SOUP \citep{Zhu2019} are developed for scRNA-seq data, and \tet{chrom}VAR \citep{Schep2017}, \tet{sc}ABC \citep{Zama2018}, SCALE \citep{Lei2019} and \tet{cis}Topic \citep{Bravo2019} are developed for scATAC-seq data. A more comprehensive discussion is presented in \citet{Lin2019}. Some methods are developed for the integrative analysis of single-cell genomic data, including Seurat \citep{Butler2018, Butler2019}, MOFA \citep{Arge2018}, \tet{couple}NMF \citep{Duren2018}, scVDMC \citep{Zhang2018}, Harmony \citep{Kors2019}, \tet{sc}ACE\citep{Lin2019} and MOFA+ \citep{Arge2020}. \citet{David2020} presented a more comprehensive discussion on integration of single-cell data across samples, experiments, and types of measurement. In real-world applications, for instance, we may better cluster scATAC-seq data by using the knowledge from scRNA-seq data, or better cluster scRNA-seq data from mouse by inference from human data. This raises a critical question on how can we apply knowledge learned from one dataset in one domain to cluster another dataset from a different domain.

In this paper, we focus on the problem of clustering single-cell genomic data across different domains. For example, we may typically have an auxiliary unlabeled dataset $\tc{A}$ from one domain (say, scRNA-seq data from human), and a target unlabeled dataset $\tc{T}$ from a different domain (say, scRNA-seq data from mouse). The two datasets follow different distributions. Target data $\tc{T}$ may consist of a collection of unlabeled data from which it is hard to learn a good feature representation - clustering directly on $\tc{T}$ may therefore perform poorly. It may be easier to learn a good feature representation from auxiliary data $\tc{A}$, which can be due to its larger sample size or less noise than $\tc{T}$. Therefore, incorporating the auxiliary data $\tc{A}$, we may achieve better clustering on the target data $\tc{T}$. This problem falls in the context of transfer learning, which utilizes knowledge obtained from one learning task to improve the performance of another \citep{Caru1997, Pan2009}, and it can be considered as an instance of unsupervised transfer learning \citep{Teh2006}, since all of the data are unlabeled.
\begin{figure}[ht]
\begin{center}
\centerline{\includegraphics[width=0.7\columnwidth]{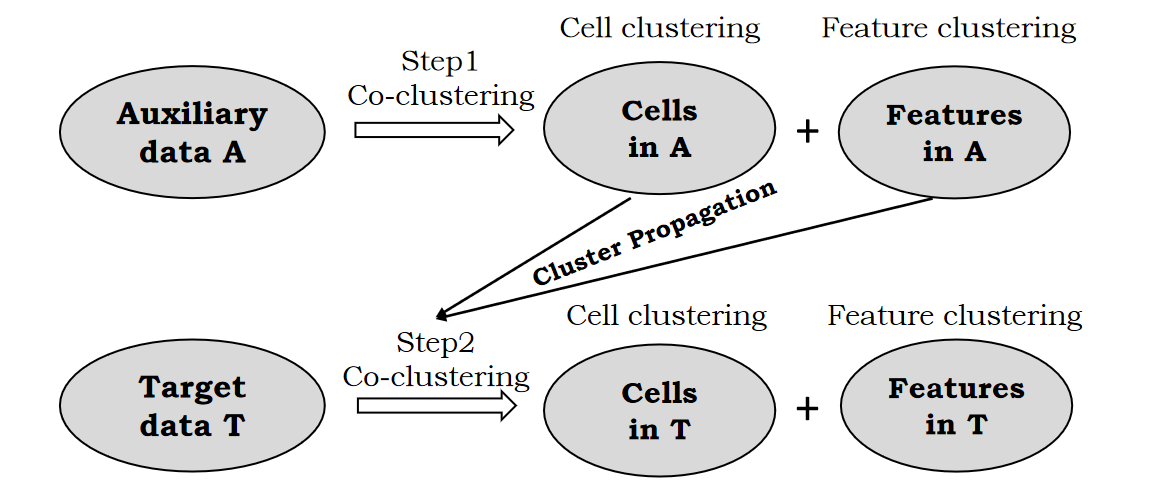}}
\caption{The model of our elastic coupled co-clustering.}
\label{fig:model}
\end{center}
\end{figure}

In this work, we propose a novel co-clustering-based transfer learning model to address this problem. A schematic plot of our proposed model is shown in Figure~\ref{fig:model}. Auxiliary data $\tc{A}$ and target data $\tc{T}$ can be regarded as two matrices with cells in the rows and genomic features in the columns. The co-clustering framework \citep{Dhillon2003}, which clusters cells and features simultaneously, is utilized in this work. Our proposed approach is composed of two steps: in Step 1, we co-cluster auxiliary data $\tc{A}$ and obtain the optimal clustering results for cells and features; in Step 2, we co-cluster target data $\tc{T}$ by transferring knowledge from the clusters of cells and features learned from $\tc{A}$. The degree of cluster propagation is elastically controlled by learning adaptively from the data, and we refer to our model as elastic coupled co-clustering (\tet{elastic}C3). If auxiliary data $\tc{A}$ and target data $\tc{T}$ are highly related, the degree of cluster propagation will be higher. On the contrary, if $\tc{A}$ and $\tc{T}$ are less related, the degree of knowledge transfer will be lower. The contributions of this paper are as follows:
\begin{itemize}
\item To the best of our knowledge, this is the first work introducing unsupervised transfer learning for clustering single-cell genomic data across different data types.
\item To ensure wide application, the model proposed in this paper can elastically control the degree of knowledge transfer and is applicable when cluster numbers of cells in auxiliary data and target data are different.
\item Our algorithm significantly boosts clustering performance on single-cell genomic data over traditional learning algorithms.
\end{itemize}

\section{Problem Formulation}
\label{sec:PF}

\begin{figure*}[ht]
\begin{center}
\centerline{\includegraphics[width=\columnwidth]{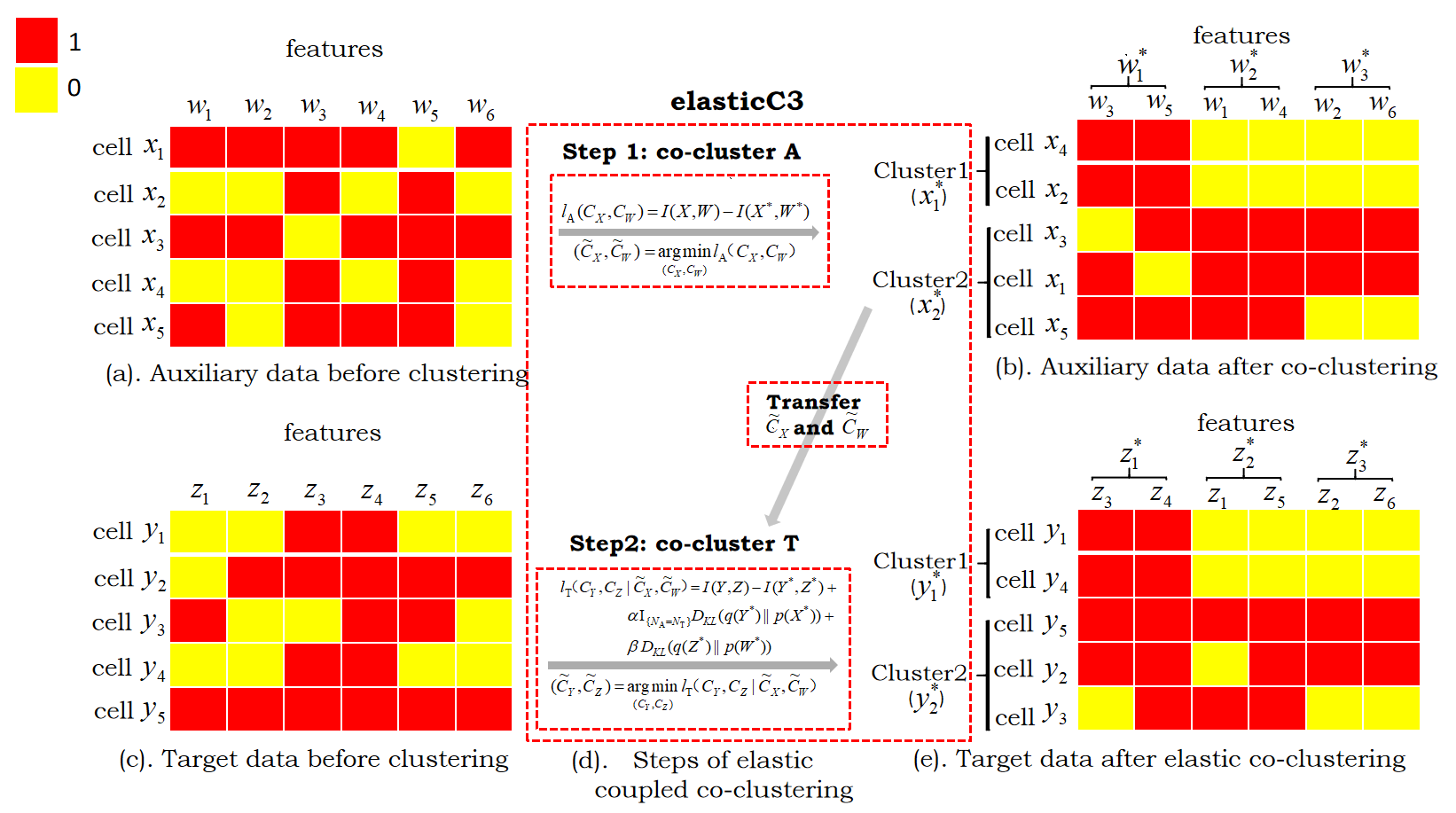}}
\caption{Toy example of elastic coupled co-clustering for single-cell genomic data. Red color means active and yellow color means inactive. $n_{\tc{A}}=n_{\tc{T}}=5,k=6,N_{\tc{A}}=N_{\tc{T}}=2,K=3$.}
\label{fig:idea}
\end{center}
\end{figure*}

We first use a toy example (Figure \ref{fig:idea}) to illustrate our method. We regard two $5 \times 6$ matrices as the auxiliary data (denoted as $\tc{A}$) and the target data (denoted as $\tc{T}$), respectively, with all matrix elements being either 1 or 0. Let $X$ and $Y$ be discrete random variables taking values from the sets of cell indexes $\{x_{1},...,x_{n_{\tc{A}}}\}$ in the auxiliary data $\tc{A}$ and $\{y_{1},...,y_{n_{\tc{T}}}\}$ in the target data $\tc{T}$, respectively. Let $W$ and $Z$ be discrete random variables for the respective feature spaces of these data, taking values from the sets of feature indexes $\{w_{1},...,w_{k}\}$ and $\{z_{1},...,z_{k}\}$. Take the auxiliary data in Figure \ref{fig:idea}(a) as an example: $X=x_{i}, i = 1,\dots, 5$ means the selected cell is the $i$-th cell among the $5$ cells and $W=w_{j}, j = 1,\dots,6$ means the selected feature is the $j$-th feature among the $6$ features. The meanings of $Y$ and $Z$ (Figure \ref{fig:idea}(c)) are the same as that for $X$ and $W$.

Let $p(X,W)$ be the joint probability distribution for $X$ and $W$, which can be represented by an $n_{\tc{A}} \times k$ matrix. $p(X=x_{i}, W = w_{j})$ represents the probability of the $j$-th gene being active: the $j$-th gene is expressed in scRNA-seq data or the $j$-th genomic region is accessible in scATAC-seq data in the $i$-th cell. The probability is estimated from the observed auxiliary data \tc{A}, and we have
$
p(X=x_{i}, W = w_{j})=\frac{\tc{A}_{ij}}{\sum_{u=1}^{n_{\tc{A}}}\sum_{v=1}^{k}{\tc{A}_{uv}}},
$
where the $\tc{A}_{uv}$ are elements of auxiliary data observations: $\tc{A}_{uv} = 1 $ if the $v$-th feature is active in the $u$-th cell, and $\tc{A}_{uv} = 0$ otherwise. The marginal probability distributions are then expressed as $p(X=x_{i})=\frac{\sum_{j=1}^{k}\tc{A}_{ij}}{\sum_{u=1}^{n_{\tc{A}}}\sum_{v=1}^{k}{\tc{A}_{uv}}}$ and $p(W=w_{j})=\frac{\sum_{i=1}^{n_{\tc{A}}}\tc{A}_{ij}}{\sum_{u=1}^{n_{\tc{A}}}\sum_{v=1}^{k}{\tc{A}_{uv}}}$. For the target data, $\tc{T}=(\tc{T}_{ij})_{n_{\tc{T}} \times k}$ is the observed data matrix. $q(Y,Z)$ is the joint probability distribution for $Y$ and $Z$. $q(Y)$ and $q(Z)$ are the marginal probabilities calculated similarly to that in the auxiliary data.

Our goal is to group similar cells and features into clusters (Figure~\ref{fig:idea}(b) and Figure~\ref{fig:idea}(e)). Suppose we want to cluster the cells in auxiliary data $\tc{A}$ and target data $\tc{T}$ into $N_{\tc{A}}$ and $N_{\tc{T}}$ clusters \footnote{Often the number of clusters is unknown, and we usually combine other exploratory analysis including visualization with clustering to determine the number of clusters in practice.}correspondingly, and cluster the features in $\tc{A}$ and $\tc{T}$ into $K$ clusters. Let $X^{\ast}$ and $Y^{\ast}$ be discrete random variables that take values from the sets of cell cluster indexes $\{x_{1}^{\ast},...,x_{N_{\tc{A}}}^{\ast}\}$ and $\{y_{1}^{\ast},...,y_{N_{\tc{T}}}^{\ast}\}$, respectively. Let $W^{\ast}$ and $Z^{\ast}$ be discrete random variables that take values from the sets of feature cluster indexes $\{w_{1}^{\ast},...,w_{K}^{\ast}\}$ and $\{z_{1}^{\ast},...,z_{K}^{\ast}\}$, respectively. We use $C_{X}(\cdot)$ and $C_{W}(\cdot)$ to represent the clustering functions for auxiliary data and $C_{X}(x)=x_{i}^{\ast}$ $(i=1,...,N_{\tc{A}})$ indicates that cell $x$ belongs to cluster $x_{i}^{\ast}$ and $C_{W}(w)=w_{i}^{\ast}$ $(i=1,...,K)$ indicates that feature $w$ belongs to cluster $w_{i}^{\ast}$. For the target data, the clustering functions $C_{Y}(\cdot)$ and $C_{Z}(\cdot)$ are defined in the same way as that for the auxiliary data. The tuples $(C_{X}, C_{W})$ and $(C_{Y}, C_{Z})$ are referred to as co-clustering \citep{Dhillon2003}.

Let $p(X^{\ast},W^{\ast})$ be the joint probability distribution of $X^{\ast}$ and $W^{\ast}$, which can be represented as an $N_{\tc{A}} \times K$ matrix. This distribution can be expressed as
\begin{equation}
\label{equa:dis2}
p(X^{\ast}=x^{\ast}_{i}, W^{\ast} = w^{\ast}_{j}) = \sum_{x \in \{C_{X}(x) = x^{\ast}_{i}\}}\sum_{w \in \{C_{W}(w)=w^{\ast}_{j}\}}p(X=x,W=w).
\end{equation}
The marginal probability distributions are then expressed as $p(X^{\ast}=x^{\ast}_{i})= \sum_{j=1}^{K}p(X^{\ast}=x^{\ast}_{i}, W^{\ast} = w^{\ast}_{j})$ and $p(W^{\ast} = w^{\ast}_{j})=\sum_{i=1}^{N_{\tc{A}}}p(X^{\ast}=x^{\ast}_{i}, W^{\ast} = w^{\ast}_{j})$. For the target data, $q(Y^{\ast}, Z^{\ast})$, $q(Y^{\ast})$ and $q(Z^{\ast})$ are defined and calculated similarly to those for the auxiliary data.

The goal of elastic coupled co-clustering in this work is to find the optimal cell clustering function $C_{Y}$ on the target data $\tc{T}$ by co-clustering the target data and utilizing the information of $(C_{X}, C_{W})$ learned from auxiliary data $\tc{A}$ (Figure \ref{fig:idea}(d)).

\section{Elastic Coupled Co-clustering Algorithm}
\label{sec:ECC}
In this section, we first present our elastic coupled co-clustering (\tet{elastic}C3) algorithm, and then discuss its theoretical properties.

\subsection{Objective Function}
Based on the information theoretic co-clustering \citep{Dhillon2003}, the objective function of co-clustering between instances and features is defined as minimizing the loss in mutual information after co-clustering. For auxiliary data $\tc{A}$, the objective function of co-clustering can be expressed:
\begin{equation}
\label{equa:obj1}
\ell_{\tc{A}}(C_{X},C_{W}) = I(X,W)-I(X^{\ast},W^{\ast}),
\end{equation}
where $I(\cdot,\cdot)$ denotes the mutual information between two random variables: \\$I(C,D) = \sum_{c \in C}\sum_{d \in D}g(c,d)\text{log}\frac{g(c,d)}{g(c)g(d)}$\citep{Cover1991}. In practice, we only consider the elements satisfying $g(c,d)\neq 0$.
We propose the following objective function for elastic coupled co-clustering of the target data $\tc{T}$:
\begin{equation}
\label{equa:obj2}
\begin{split}
\ell_{\tc{T}}(C_{Y},C_{Z}|\tilde{C}_{X},\tilde{C}_{W}) = &I(Y,Z)-I(Y^{\ast},Z^{\ast}) + \alpha \mathds{1}_{\{N_{\tc{A}}=N_{\tc{T}}\}}D_{\tc{KL}}(q(Y^{\ast})||p(X^{\ast}))+\\
&\beta D_{\tc{KL}}(q(Z^{\ast})||p(W^{\ast})),
\end{split}
\end{equation}
where $\mathds{1}_{\{N_{\tc{A}}=N_{\tc{T}}\}}$ equals to 1 if $N_{\tc{A}}=N_{\tc{T}}$ and 0 otherwise, $(\tilde{C}_{X},\tilde{C}_{W}) = \Min_{C_{X},C_{W}}\ell_{\tc{A}}(C_{X},C_{W})$, and $D_{\tc{KL}}(\cdot || \cdot)$ denotes the Kullback-Leibler divergence between two probability distributions \citep{Cover1991}, where $D_{\tc{KL}}(g||h)= \sum_{x}g(x)\text{log}\frac{g(x)}{h(x)}$. The term $I(Y,Z)-I(Y^{\ast},Z^{\ast})$ measures the loss in mutual information after co-clustering for the target data $\tc{T}$. $D_{\tc{KL}}(q(Y^{\ast})||p(X^{\ast}))$ and $D_{\tc{KL}}(q(Z^{\ast})||p(W^{\ast}))$ are two distribution-matching terms between $\tc{A}$ and $\tc{T}$ after co-clustering, in terms of the row-dimension (cells) and the column-dimension (features), respectively. When the numbers of cell clusters are different between auxiliary data \tc{A} and target data \tc{T} ($N_\tc{A} \neq N_\tc{T}$), the distribution-matching term for the row-dimension disappears. $\alpha$ and $\beta$ are non-negative hyper-parameters that elastically control how much information should be transferred, and both tend to be higher if the auxiliary data are more similar to the target data. How the parameters $\alpha$ and $\beta$ are tuned is discussed in the following section.

\subsection{Optimization}
The optimization for the objective function in Equation (\ref{equa:obj2}) can be divided into two separate steps. In Step 1, as shown at the top of Figure \ref{fig:idea}(d), we use the co-clustering algorithm by \citet{Dhillon2003} to solve the following optimization problem:
\begin{equation}
\label{equa:step1}
(\tilde{C}_{X},\tilde{C}_{W}) = \Min_{(C_{X},C_{W})}\ell_{\tc{A}}(C_{X},C_{W}).
\end{equation}
Details are given in \ref{ap:algo}. In Step 2, we transfer the estimated $\tilde{C}_{X}$ and $\tilde{C}_{W}$ from Step 1, as shown at the bottom of Figure \ref{fig:idea}(d), to solve the following optimization problem:
\begin{equation}
\label{equa:obj3}
(\tilde{C}_{Y},\tilde{C}_{Z}) = \Min_{(C_{Y},C_{Z})}\ell_{\tc{T}}(C_{Y},C_{Z}|\tilde{C}_{X},\tilde{C}_{W}).
\end{equation}
We first rewrite the term $I(Y,Z)-I(Y^{\ast},Z^{\ast})$ in Equation (\ref{equa:obj2}) in a similar manner as rewriting $I(X,W)-I(X^{\ast},W^{\ast})$ in Equations (\ref{equa:opt0}), (\ref{equa:opt1}) and (\ref{equa:opt2}) in \ref{ap:algo}, and we have
\begin{equation}
\label{equa:opt3}
\begin{split} D_{\tc{KL}}(q(Y,Z)||q^{\ast}(Y,Z))  & =\sum_{y^{\ast} \in \{y_{1}^{\ast},...,y_{N_{\tc{T}}}^{\ast}\}}\sum_{y \in \{y: C_{Y}(y) = y^{\ast}\}}q(y)D_{\tc{KL}}(q(Z|y)||q^{\ast}(Z|y^{\ast}))\\
                          & =\sum_{z^{\ast} \in \{z_{1}^{\ast},...,z_{K}^{\ast}\}}\sum_{z \in \{z: C_{Z}(z) = z^{\ast}\}}q(z)D_{\tc{KL}}(q(Y|z)||q^{\ast}(Y|z^{\ast})),
\end{split}
\end{equation}
where $q^{\ast}(z|y^{\ast}) \triangleq \frac{q^{\ast}(y,z)}{q(y)} =  \frac{q(y^{\ast},z^{\ast})}{q(y^{\ast})}\frac{q(z)}{q(z^{\ast})}$ and $q^{\ast}(y|z^{\ast}) \triangleq \frac{q^{\ast}(y,z)}{q(z)} =  \frac{q(y^{\ast},z^{\ast})}{q(z^{\ast})}\frac{q(y)}{q(y^{\ast})}$.

We can iteratively update $C_{Y}$ and $C_{Z}$ to minimize $\ell_{\tc{T}}$. First, given $C_{Z}$, minimizing $\ell_{\tc{T}}$ is equivalent to minimizing $\sum_{y^{\ast} \in \{y_{1}^{\ast},...,y_{N_{\tc{T}}}^{\ast}\}}\sum_{y \in \{y: C_{Y}(y) = y^{\ast}\}}q(y)Q_{\alpha}(Y^{\ast}|C_{Z},\tilde{C}_{X})$,
where
\[
Q_{\alpha}(Y^{\ast}|C_{Z},\tilde{C}_{X}) \triangleq D_{\tc{KL}}(q(Z|y)||q^{\ast}(Z|y^{\ast}))+\frac{\alpha D_{\tc{KL}}(q(Y^{\ast})||p(X^{\ast}))}{n_{\tc{T}}N_{\tc{T}}q(y)}\mathds{1}_{\{N_{\tc{A}}=N_{\tc{T}}\}}.
\]
We iteratively update the cluster assignment $y^{\ast}$ for each cell $y$ in the target data, fixing the cluster assignment for the other cells:
\begin{equation}
\label{equa:cy}
C_{Y}(y) = \Min_{y^{\ast} \in \{y_{1}^{\ast},...,y_{N_{\tc{T}}}^{\ast}\}} Q_{\alpha}(Y^{\ast}|C_{Z},\tilde{C}_{X}).
\end{equation}
Second, given $C_{Y}$, minimizing $\ell_{\tc{T}}$ is equivalent to minimizing\\
$
\sum_{z^{\ast} \in \{z_{1}^{\ast},...,z_{K}^{\ast}\}}\sum_{z \in \{z: C_{Z}(z) = z^{\ast}\}}q(z)R_{\beta}(Z^{\ast}|C_{Y},\tilde{C}_{W}),
$
where
\[
R_{\beta}(Z^{\ast}|C_{Y},\tilde{C}_{W}) \triangleq D_{\tc{KL}}(q(Y|z)||q^{\ast}(Y|z^{\ast}))+\frac{\beta D_{\tc{KL}}(q(Z^{\ast})||p(W^{\ast}))}{n_{\tc{T}}N_{\tc{T}}q(z)}.
\]
We iteratively update the cluster assignment $z^{\ast}$ for each feature $z$ in the target data, fixing the cluster assignment for the other features:
\begin{equation}
\label{equa:cz}
C_{Z}(z) = \Min_{z^{\ast} \in \{z_{1}^{\ast},...,z_{K}^{\ast}\}} R_{\beta}(Z^{\ast}|C_{Y},\tilde{C}_{W}).
\end{equation}

Summaries of Steps 1 and 2 are given in Algorithm~\ref{algorithm}. Note that our model has three hyper-parameters: the non-negative $\alpha$ and $\beta$, and K (the number of clusters in feature spaces W and Z). We perform grid-search to choose the optimal combination of parameters, and we will show grid search performs well in both simulated data and real data in Section \ref{sec:experiments}.

\begin{algorithm}[H]
\DontPrintSemicolon
  \KwInput{$\tc{A}$,$\tc{T}$,$N_{\tc{A}}$,$N_{\tc{T}}$,$\alpha$,$\beta$,$K$ and the number of iterations $I_{\tc{A}}$ and $I_{\tc{T}}$}
  \KwOutput{$C_{Y}^{(I_{\tc{T}})}$}
  \tcc{Initialization}
  Calculate $p$ and $q$, and initialize $C_{X}^{(0)}$, $C_{W}^{(0)}$, $C_{Y}^{(0)}$ and $C_{Z}^{(0)}$;\\
  Initialize $p^{\ast(0)}$ based on $p$, $C_{X}^{(0)}$, $C_{W}^{(0)}$ and initialize $q^{\ast(0)}$ based on $q$, $C_{Y}^{(0)}$, $C_{Z}^{(0)}$.\\
  \tcc{Step 1: details in Appendix 1}
    \While{$i < I_{\tc{A}}$}
   {
   Update $C_{X}^{(i)}$ based on $p$, $p^{\ast(i-1)}$ and Equation (\ref{equa:cx}).\\
   Update $C_{W}^{(i)}$ based on $p$, $p^{\ast(i-1)}$ and Equation (\ref{equa:cw}).\\
   Update $p^{\ast(i)}$ based on $C_{X}^{(i)}$, $C_{W}^{(i)}$ and Equation (\ref{equa:opt1}). 
   }
  Return $C_{X}^{(\tc{I}_{\tc{A}})}$ and $C_{W}^{(\tc{I}_{\tc{A}})}$ as the final clustering functions for the auxiliary data $\tc{A}$.\\
  \tcc{Step 2}
  \While{$i < I_{\tc{T}}$}
   {
   Update $C_{Y}^{(i)}$ based on $q$, $q^{\ast(i-1)}$,$C_{X}^{(\tc{I}_{\tc{A}})}$ and Equation (\ref{equa:cy}).\\
   Update $C_{Z}^{(i)}$ based on $q$, $q^{\ast(i-1)}$,$C_{W}^{(\tc{I}_{\tc{A}})}$ and Equation (\ref{equa:cz}).\\
   Update $q^{\ast(i)}$ based on $C_{Y}^{(i)}$, $C_{Z}^{(i)}$.
   }
  Return $C_{Y}^{(\tc{I}_{\tc{T}})}$ as the final clustering functions for the target data $\tc{T}$.
\caption{The \tet{elastic}C3 Algorithm}
\label{algorithm}
\end{algorithm}

\subsection{Theoretical Properties}
First, we give the monotonically decreasing property of the objective function of the \tet{elastic}C3 algorithm in the following theorem:
\begin{theorem}
Let the value of objective function $\ell_{\tc{T}}$ in the $i$-th iteration be
\begin{equation}
\begin{split}
\ell_{\tc{T}}(C_{Y}^{\ast(i)},C_{Z}^{\ast(i)}|\tilde{C}_{X},\tilde{C}_{W}) &= I(Y,Z)-I(Y^{\ast(i)},Z^{\ast(i)}) + \qquad \\\alpha \mathds{1}_{\{N_{\tc{A}}=N_{\tc{T}}\}} D_{\tc{KL}}(q(Y^{\ast(i)})||p(X^{\ast}))
&+\beta D_{\tc{KL}}(q(Z^{\ast(i)})||p(W^{\ast})).
\end{split}
\end{equation}
Then, we have
\begin{equation}
\ell_{\tc{T}}(C_{Y}^{\ast(i)},C_{Z}^{\ast(i)}|\tilde{C}_{X},\tilde{C}_{W}) \geq \ell_{\tc{T}}(C_{Y}^{\ast(i+1)},C_{Z}^{\ast(i+1)}|\tilde{C}_{X},\tilde{C}_{W}).
\end{equation}
\end{theorem}
The proof of \textbf{Theorem} 1 is given in \ref{ap:proof}. Because the search space is finite and the objective function is non-increasing in the iterations, Algorithm~\ref{algorithm} converges in a finite number of iterations. Second, our proposed algorithm converges to a local minimum, since finding the global optimal solution is NP-hard.

Finally, we analyze the computational complexity of our algorithm. Suppose the total number of cell-feature co-occurrences (means 1 within data matrices) in the auxiliary dataset is $d_{\tc{A}}$ and in the target dataset is $d_{\tc{T}}$. For each iteration, updating $C_{X}$ and $C_{W}$ in Step 1 takes $\mathcal{O}((N_{\tc{A}}+K)\cdot d_{\tc{A}})$, while updating $C_{Y}$ and $C_{Z}$ in Step 2 takes $\mathcal{O}((N_{\tc{T}}+K)\cdot d_{\tc{T}})$. The number of iterations is $\tc{I}_{\tc{A}}$ in Step 1 and $\tc{I}_{\tc{T}}$ in Step 2. Therefore, the time complexity of our \tet{elastic}C3 algorithm is $\mathcal{O}((N_{\tc{A}}+K)\cdot d_{\tc{A}}\cdot \tc{I}_{\tc{A}}+(N_{\tc{T}}+K)\cdot d_{\tc{T}}\cdot \tc{I}_{\tc{T}}))$. In the experiments, it is shown that $\tc{I}_{\tc{A}} = \tc{I}_{\tc{T}} = 10$ is enough for convergence in Figure \ref{fig:conv}. We may consider the number of clusters $N_{\tc{A}}$, $N_{\tc{T}}$ and $K$ as constants; thus the time complexity of \tet{elastic}C3 is $\mathcal{O}(d_{\tc{A}}+d_{\tc{T}})$. Because our algorithm needs to store all of the cell-feature co-occurrences, it has a space complexity of $\mathcal{O}(d_{\tc{A}}+d_{\tc{T}})$.

\section{Experiments}
\label{sec:experiments}
To evaluate the performance of our algorithm, we conduct experiments on simulated datasets and two single-cell genomic datasets.

\subsection{Data Sets}
To generate the simulated single-cell data, we follow the simulation setup given in \citet{Lin2019}. We set the scATAC-seq data as the auxiliary data $\tc{A}$ and set the scRNA-seq data as the target data $\tc{T}$. We set the number of clusters $N_{\tc{A}}=N_{\tc{T}}=2$. Details for generating $\tc{A}$ and $\tc{T}$ are given in the \ref{ap:gene}. We set the number of cells in both $\tc{A}$ and $\tc{T}$ as $n_{\tc{A}}=n_{\tc{T}}=100$, and set the number of features as $k=100$. We assume that only a subset of features are highly correlated in scRNA-seq and scATAC-seq data, and the other features have no more correlation than random. We vary the percentage of the highly correlated features (percentage = 0.1, 0.5, and 0.9)(details the first column in Table \ref{tab:simu1}).

Our experiment also contains two real single-cell genomic datasets. Real data 1 includes human scRNA-seq data as the auxiliary data and human scATAC-seq data as the target data, where 233 K562 and 91 HL60 scATAC-seq cells are obtained from \citet{Buenr2015}, and 42 K562 and 54 HL60 deeply sequenced scRNA-seq cells are obtained from \citet{Pollen2014}. True cell labels are used as a benchmark for evaluating the performance of the clustering methods. Real data 2 includes human scRNA-seq data as the auxiliary data and mouse scRNA-seq data as the target data. There are three cell types and the datasets are downloaded from \tet{panglaodb.se} \citep{Fran2019}. For the mouse data, 179 pulmonary alveolar type 2 cells, 99 clara cells and 14 ependymal cells are obtained under the accession number SRS4237518. For the human data, 193 pulmonary alveolar type 2 cells, 113 clara cells and 58 ependymal cells are obtained under accession number SRS4660846. We use the cell-type annotation \citep{Ange2019} as a benchmark for evaluating the performance of the clustering methods. For each real data, we also consider the setting where $N_{\tc{A}} \neq N_{\tc{T}}$ by removing one cell type from the auxiliary data, referred as Setting 2 (details in the first column in Table \ref{tab:real1}).

\subsection{Experimental Results}
For the simulation analysis, we compare our method with the classic unsupervised transfer clustering method STC \citep{Dai2008} and $k$-means clustering. For real data analysis, we implement variable selection before performing clustering, and select 100 most variable features for each real dataset (\ref{ap:feature}). After variable selection, we binarize the data matrices by setting the non-zero entries to 1. For real data analysis, we compare our method with STC \citep{Dai2008}, co-clustering \citep{Dhillon2003} and two commonly used clustering methods for single-cell genomic data, including SC3 \citep{Kise2017} and SIMLR \citep{Wang2017}. \footnote{For our method \tet{elasticC3}, we use grid search to tune the hyper-parameters $\alpha$, $\beta$ and K . We choose the search domains $\alpha \in [0,1]$, $\beta \in [0,1]$ and $K \in (0,10)$. The methods STC and co-clustering require input to be binary. Both SC3 and SIMLR use log2(TPM+1) as input for the real scRNA-seq data. The hyper-parameters in STC and co-clustering are also tuned by grid search, as presented in their original publication.} Note that the method co-clustering is the same as implementing \tet{elastic}C3 with $\alpha=0$ and $\beta=0$, and no knowledge is transferred between auxiliary data and target data.  We use four criteria to evaluate the clustering results, including normalized mutual information (NMI), adjusted Rand index (ARI), Rand index (RI) and purity.

\begin{table}[ht!]
\centering
\caption{Clustering results for simulated scRNA-seq data (target data) over 50 independent runs. The values of the hyper-parameters $\alpha$,$\beta$ and $K$ for \tet{elastic}C3 algorithm are shown, where they are tuned by grid search.}
\begin{tabular}{ccccccccccc}
\toprule
Simulated Data Setting & &Methods&    &NMI& &ARI& &RI& &purity\\
\midrule
   Setting 1                  &&$\tet{elastic}$C3&    &0.214& &0.262& &0.631&&0.752\\
 $\text{percentage} =0.1$       &&STC& &0.142& &0.172& &0.586& &0.701\\
$\alpha=0.1, \beta = 0, K = 3$          &&$k$-means& &0.116& &0.136& &0.568& &0.671\\
\midrule
  Setting 2                           &&$\tet{elastic}$C3& &0.282& &0.345& &0.674& &0.796\\
$\text{percentage} =0.5$                         &&STC& &0.208& &0.255& &0.628& &0.753\\
$\alpha=0.5, \beta = 0.01, K = 3$       &&$k$-means& &0.229& &0.263& &0.632& &0.748\\
\midrule
  Setting 3                            &&$\tet{elastic}$C3& &0.379& &0.454& &0.727& &0.837\\
$\text{percentage}=0.9$                           &&STC& &0.374& &0.440& &0.720& &0.830\\
$\alpha=0.9, \beta = 0.04, K = 3$      &&$k$-means& &0.211& &0.241& &0.621& &0.735\\
\bottomrule
\end{tabular}%
\label{tab:simu1}
\end{table}

\begin{table}[ht!]
\centering
\caption{Clustering results for the target data in real datasets. By doing grid search, $\alpha=0.1, \beta = 0.1, K=3$ on real data 1 and $\alpha=0.05, \beta = 0.01, K=3$ on real data 2 for our elasticC3 method.}
\resizebox{\columnwidth}{!}{%
\begin{tabular}{ccccccccccc}
\toprule
Real Data Sets & &Methods& &NMI& &ARI& &RI& &purity \\
\midrule
    Real data 1                         &&$\tet{elastic}$C3 (Setting1)& &0.610& &0.743& &0.875& &0.933\\
$\tc{A}$: human scRNA-seq data          &&$\tet{elastic}$C3 (Setting2)& &0.535& &0.658& &0.832& &0.908\\
$\tc{T}$: human scATAC-seq data         &&STC (Setting1)& &0.495& &0.616& &0.811& &0.895\\
Setting 1: complete data                &&STC (Setting2)& &0.472& &0.587& &0.796& &0.885\\
($N_{\tc{A}}=N_{\tc{T}}=2$)             &&Co-clustering& &0.502& &0.617& &0.811& &0.895\\
Setting 2: K562 removed from $\tc{A}$     &&SC3& &0.000& &0.001& &0.504& &0.710\\
($N_{\tc{A}}=1,N_{\tc{T}}=2$)           &&SIMLR& &0.025& &0.046& &0.525& &0.710\\
\midrule
   Real data 2                         &&$\tet{elastic}$C3 (Setting1)& &0.564& &0.682& &0.841& &0.880\\
$\tc{A}$: human scRNA-seq data         &&$\tet{elastic}$C3 (Setting2)& &0.515& &0.629& &0.815& &0.863\\
$\tc{T}$: mouse scRNA-seq data         &&STC (Setting1)& &0.454& &0.434& &0.718& &0.836\\
Setting 1: complete data              &&STC (Setting2)& &0.397& &0.425& &0.714& &0.788\\
   ($N_{\tc{A}}=N_{\tc{T}}=3$)         &&Co-clustering& &0.510& &0.625& &0.813& &0.860\\
Setting 2: alveolar type 2 removed from $\tc{A}$ &&SC3& &0.378& &0.318& &0.661& &0.849\\
   ($N_{\tc{A}}=2,N_{\tc{T}}=3$)       &&SIMLR& &0.405& &0.244& &0.618& &0.709\\
\bottomrule
\end{tabular}%
}
\label{tab:real1}
\end{table}

\subsubsection{Performance}
We present in Table \ref{tab:simu1} the clustering results for the simulated scRNA-seq data (target data $\tc{T}$). Across different settings, the trends are similar for the four clustering criteria, including NMI, ARI, RI and purity. When the percentage of highly correlated features increases from 0.1 to 0.9, more information is shared among the feature space W for the auxiliary data and the feature space Z for the target data. The clustering results for $\tet{elastic}$C3 and STC are improved, because they can transfer more informative knowledge from the auxiliary data to improve clustering of the target data. When the features W and Z are less similar (percentage = 0.1 and 0.5), STC does not perform as well as our method \tet{elastic}C3, because STC assumes that W and Z are the same, while \tet{elastic}C3 assumes that W and Z are different and elastically controls the degree of knowledge transfer by introducing the term $\beta D_{\tc{KL}}(q(Z^{\ast})||p(W^{\ast}))$ in Equation (\ref{equa:obj2}). When W and Z are more similar (percentage = 0.9), STC is comparable to \tet{elastic}C3. We also note that our algorithm $\tet{elastic}$C3 can adaptively learn the degree of knowledge transferring from the data because the parameters $\alpha$ and $\beta$ increase as the similarity of the auxiliary data and target data increases. In Setting 1, the features W and Z are less related and the tuning parameter $\beta=0$ in \tet{elastic}C3. Finally, because $k$-means clustering cannot transfer knowledge from auxiliary data to target data, it does not work as well as $\tet{elastic}$C3 and STC.

The clustering results for the target data in two real single-cell genomic datasets are shown in Table 2. We see that $\tet{elastic}$C3 performs better in Setting 1 than in Setting 2 in the two datasets, because one cell type is removed from the auxiliary data in Setting 2, thereby losing information that can be transferred. Knowledge transfer improves clustering of the target data as $\tet{elastic}$C3 outperforms the methods that do not incorporate knowledge transfer, including co-clustering, SC3 and SIMLR in both datasets. The distributions of the features are likely different in the auxiliary data and target data: in real data 1, the distribution of gene expression is likely different from the distribution of promoter accessibility; in real data 2, the distributions of gene expression may be different between human and mouse. STC treats the features in the auxiliary data and target data as the same and performs knowledge transfer. In both datasets, the performance of STC is not as good as that of $\tet{elastic}$C3, and it is even worse than that of co-clustering, especially for real data 2. The degree of shared information is likely lower in real data 2, as indicated by the smaller values in $\alpha$ and $\beta$ when we implement elasticC3. In summary, $\tet{elastic}$C3 adaptively learns the degree of knowledge transfer in auxiliary data and target data, and it outperforms methods that do not allow for knowledge transfer (co-clustering, SC3 and SIMLR) and methods that do not control the degree of knowledge transfer (STC). In addition, $\tet{elastic}$C3 works well even when the number of cell clusters differ between auxiliary data and target data, ensuring its wider application.

\subsubsection{Convergence}
\begin{figure}[ht]
\begin{center}
\centerline{\includegraphics[width=0.7\columnwidth]{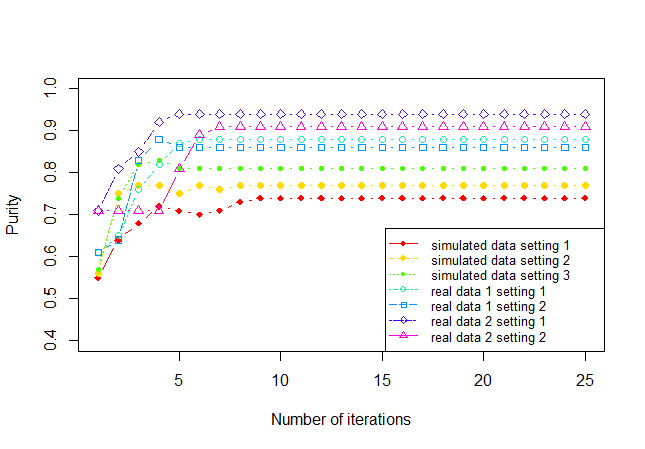}}
\caption{The purity curves after each iteration.}
\label{fig:conv}
\end{center}
\end{figure}
We have proven the convergence of $\tet{elastic}$C3 in Theorem 1, and now we show its convergence property empirically. Figure \ref{fig:conv} shows the curve of purity vs number of iterations on the seven datasets (For simulated data, we randomly select one run for each setting). Our algorithm $\tet{elastic}$C3 converges within 10 iterations. Using the other three clustering criteria gives similar convergence results.

\section{Conclusion and Future Work}
In this paper, we have developed $\tet{elastic}$C3 for the integrative analysis of multiple single-cell genomic datasets. Our proposed method, $\tet{elastic}$C3, was developed under the unsupervised transfer learning framework, where the knowledge learned from an auxiliary data is utilized to improve the clustering results of target data. Our algorithm consists of two separate steps. In Step 1, we cluster both the cells and features (i.e. co-cluster) in the auxiliary data, and in Step 2 we co-cluster the target data by elastically transferring the knowledge learned in Step 1. We prove the convergence of $\tet{elastic}$C3 to a local optimum. Our algorithm outperforms other commonly used clustering methods in single-cell genomics. Because the framework of \tet{elastic}C3 is general, we plan to explore its application to other areas, including text mining.

\section*{Broader Impact}
This work does not present any foreseeable societal consequences and ethical issues.

\section*{Acknowledgement}
This work has been supported by the Chinese University of Hong Kong direct grant 2018/2019 No. 4053360, the Chinese University of Hong Kong startup grant No. 4930181, and Hong Kong Research Grant Council Grant ECS No. CUHK 24301419.

\bibliography{elasticC3}
\bibliographystyle{apalike}
\nocite{Cou90}

\newpage
\begin{appendices}
\section{Details of Step 1 in elasticC3 algorithm}
\label{ap:algo}
The optimization problem \[(\tilde{C}_{X},\tilde{C}_{W}) = \Min_{(C_{X},C_{W})}\ell_{\tc{A}}(C_{X},C_{W})\] is non-convex and challenging to solve. We rewrite this objective function in the form of KL divergence, since the reformulated objective function is easier to optimize. To be more specific,
\begin{equation}
\label{equa:opt0}
\ell_{\tc{A}}(C_{X},C_{W}) = D_{\tc{KL}}(p(X,W)||p^{\ast}(X,W)),
\end{equation}
where $p^{\ast}(X,W)$ is expressed as
\begin{equation}
\label{equa:opt1}
p^{\ast}(x,w)=p(x^{\ast},w^{\ast})\frac{p(x)}{p(x^{\ast})}\frac{p(w)}{p(w^{\ast})}.
\end{equation}
The distributions $p(x^{\ast},w^{\ast})$, $p(x)$, $p(x^{\ast})$, $p(w)$, $p(w^{\ast})$ are estimated as in Section \ref{sec:PF} in the text. Further, we have
\begin{equation}
\label{equa:opt2}
\begin{split} D_{\tc{KL}}(p(X,W)||p^{\ast}(X,W))  & =\sum_{x^{\ast} \in \{x_{1}^{\ast},...,x_{N_{\tc{A}}}^{\ast}\}}\sum_{x \in \{x: C_{X}(x) = x^{\ast}\}}p(x)D_{\tc{KL}}(p(W|x)||p^{\ast}(W|x^{\ast}))\\
                          & =\sum_{w^{\ast} \in \{w_{1}^{\ast},...,w_{K}^{\ast}\}}\sum_{w \in \{w: C_{W}(w) = w^{\ast}\}}p(w)D_{\tc{KL}}(p(X|w)||p^{\ast}(X|w^{\ast})),
\end{split}
\end{equation}
where $p^{\ast}(w|x^{\ast}) \triangleq \frac{p^{\ast}(x,w)}{p(x)} =  \frac{p(x^{\ast},w^{\ast})}{p(x^{\ast})}\frac{p(w)}{p(w^{\ast})}$ and $p^{\ast}(x|w^{\ast}) \triangleq \frac{p^{\ast}(x,w)}{p(w)}  =  \frac{p(x^{\ast},w^{\ast})}{p(w^{\ast})}\frac{p(x)}{p(x^{\ast})}$. The relations in Equations (\ref{equa:opt0}) and (\ref{equa:opt2}) have been proven by \citet{Dhillon2003} and \citet{Dai2008}.

To minimize $D_{\tc{KL}}(p(X,W)||p^{\ast}(X,W))$, we can iteratively update $C_{X}$ and $C_{W}$ as follows.
\begin{itemize}
\item Fix $C_{W}$ and iteratively update the cluster assignment $x^{\ast}$ for each cell $x$ in the auxiliary data, while fixing the cluster assignments for the other cells:
\begin{equation}
\label{equa:cx}
C_{X}(x) = \Min_{x^{\ast} \in \{x_{1}^{\ast},...,x_{N_{\tc{A}}}^{\ast}\}} D_{\tc{KL}}(p(W|x)||p^{\ast}(W|x^{\ast})).
\end{equation}
\item Fix $C_{X}$ and iteratively update the cluster assignment $w^{\ast}$ for each feature $w$ in the auxiliary data, while fixing the cluster assignments for the other features:
\begin{equation}
\label{equa:cw}
C_{W}(w) = \Min_{w^{\ast} \in \{w_{1}^{\ast},...,w_{K}^{\ast}\}} D_{\tc{KL}}(p(X|w)||p^{\ast}(X|w^{\ast})).
\end{equation}
\end{itemize}

The above procedures monotonically decrease the objective function (\ref{equa:obj1})\citep{Dhillon2003}, converging to a local minimum.
\section{Proof of Theorem 1}
\label{ap:proof}
\textbf{Proof} We rewrite $\ell_{\tc{T}}(C_{Y}^{\ast(i)},C_{Z}^{\ast(i)}|\tilde{C}_{X},\tilde{C}_{W})$ as
\begin{equation*}
\begin{split}
\ell_{\tc{T}}(C_{Y}^{\ast(i)},C_{Z}^{\ast(i)}|\tilde{C}_{X},\tilde{C}_{W}) = & D_{\tc{KL}}(q(Y,Z)||q^{\ast(i)}(Y,Z))  + \alpha \mathds{1}_{\{N_{\tc{A}}=N_{\tc{T}}\}} D_{\tc{KL}}(q(Y^{\ast(i)})||p(X^{\ast}))+\\&\beta D_{\tc{KL}}(q(Z^{\ast(i)})||p(W^{\ast}))\\
                                    = & \frac{1}{2}\big(D_{\tc{KL}}(q(Y,Z)||q^{\ast(i)}(Y,Z))+2*\alpha \mathds{1}_{\{N_{\tc{A}}=N_{\tc{T}}\}} D_{\tc{KL}}(q(Y^{\ast(i)})||p(X^{\ast}))\big)+\\
                                    &\frac{1}{2}\big(D_{\tc{KL}}(q(Y,Z)||q^{\ast(i)}(Y,Z))+2*\beta D_{\tc{KL}}(q(Z^{\ast(i)})||p(W^{\ast})\big)\\
                                    = &\frac{1}{2}\sum_{y^{\ast} \in \{y_{1}^{\ast(i)},...,y_{N_{\tc{T}}}^{\ast(i)}\}}\sum_{y \in \{y: C_{Y}(y) = y^{\ast}\}}q(y)Q_{2\alpha}(Y^{\ast(i)}|C_{Z}^{\ast(i-1)},\tilde{C}_{X})+\\
                                    &\frac{1}{2} \sum_{z^{\ast} \in \{z_{1}^{\ast(i)},...,z_{N_{\tc{T}}}^{\ast(i)}\}}\sum_{z \in \{z: C_{Z}(z) = z^{\ast}\}}q(z)R_{2\beta}(Z^{\ast(i)}|C_{Y}^{\ast(i-1)},\tilde{C}_{W})\\
                                    \triangleq & \frac{1}{2}\ell_{\tc{T}1}(C_{Y}^{\ast(i)}|C_{Z}^{\ast(i-1)},\tilde{C}_{X}) + \frac{1}{2}\ell_{\tc{T}2}(C_{Z}^{\ast(i)}|C_{Y}^{\ast(i-1)},\tilde{C}_{W})\\
                                    \geq & \frac{1}{2}\ell_{\tc{T}1}(C_{Y}^{\ast(i+1)}|C_{Z}^{\ast(i)},\tilde{C}_{X}) + \frac{1}{2}\ell_{\tc{T}2}(C_{Z}^{\ast(i+1)}|C_{Y}^{\ast(i)},\tilde{C}_{W})\\
                                    =& \ell_{\tc{T}}(C_{Y}^{\ast(i+1)},C_{Z}^{\ast(i+1)}|\tilde{C}_{X},\tilde{C}_{W})
\end{split}
\end{equation*}
Note that
\[
\ell_{\tc{T}1}(C_{Y}^{\ast(i)}|C_{Z}^{\ast(i-1)},\tilde{C}_{X}) \geq  \ell_{\tc{T}1}(C_{Y}^{\ast(i+1)}|C_{Z}^{\ast(i)},\tilde{C}_{X})
\]
and
\[
\ell_{\tc{T}2}(C_{Z}^{\ast(i)}|C_{Y}^{\ast(i-1)},\tilde{C}_{W}) \geq  \ell_{\tc{T}2}(C_{Z}^{\ast(i+1)}|C_{Y}^{\ast(i)},\tilde{C}_{W})
\]
are straightforward based on Equation (\ref{equa:cy}) and (\ref{equa:cz}).

\section{Steps of data generation in simulation study}
\label{ap:gene}
Similar to the simulation scheme in \citet{Lin2019}, we generate data in simulation study as in the following:
\begin{enumerate}
\item Generate $\tf{w}^{acc}$ and $\tf{w}^{exp}$.
\[
w^{acc}_{rj}  = \left\{
                  \begin{array}{ll}
                w, \hspace{6mm} r = 1, j=1, \dots, k(1-w)/2; \\\hspace{10mm} r = 2, j=k(1-w)/2+1,\dots,k(1-w)\\
                1-w, \hspace{1mm}r = 2, j=1, \dots, k(1-w)/2; \\\hspace{10mm} r = 1, j=k(1-w)/2+1,\dots,k(1-w).
                  \end{array}
                  \right.
\]
$w^{acc}_{1j} = w^{acc}_{2j} \sim Beta(0.5,2), j = k(1-w)+1,\dots,k$.
$w^{exp}_{cj} \sim Beta(w^{acc}_{cj},10), j = 1,\dots,k(1-w)$;$w^{exp}_{1j} = w^{exp}_{2j} \sim Beta(w^{acc}_{1j},10), j = k(1-w)+1,\dots,k$.
\item Generate $z^{acc}$ and $z^{exp}$. The cluster labels are generated with equal probability 0.5.
\item Generate $u^{acc}$ and $\tilde{u}^{acc}$. $\tilde{u}^{acc}_{ij} \sim Bernoulli(0.5)$ if $u^{acc}_{ij}=1$, where $u^{acc}_{ij} \sim Bernoulli(w^{acc}_{cj})$ if $z^{acc}_{ic}=1,i=1,\dots,m$;$\tilde{u}^{acc}_{ij} = 0$ otherwise.
\item Generate $u^{exp}$ and $\tilde{v}^{exp}$. $\tilde{v}^{exp}_{lj} \sim Bernoulli(0.8)$ if $u^{exp}_{lj}=1$, where $u^{exp}_{lj} \sim Bernoulli(w^{exp}_{lj})$ if $z^{exp}_{lc}=1$;$\tilde{v}^{exp}_{lj} \sim Bernoulli(0.1)$ otherwise; $l=1,\dots,n$.
\item Generate $C$ and $G$. $C_{ij} \sim N(0, 0.6^{2})$ if $\tilde{u}^{acc}_{ij}=0$ and $C_{ij} \sim N(2, 0.6^{2})$ if $\tilde{u}^{acc}_{ij}=1$; $G_{lj} \sim N(0, \sigma^{2})$ if $\tilde{v}^{exp}_{lj}=0$ and $G_{lj} \sim N(2, \sigma^{2})$ if $\tilde{v}^{exp}_{lj}=1$.
\item Generate target data $\tc{T}$ and auxiliary data $\tc{A}$. $\tc{T}_{lj}=1$ if $\tc{G}_{lj}>0$ and $\tc{T}_{lj}=0$ otherwise; $\tc{A}_{ij}=1$ if $C_{ij}>0$ and $\tc{A}_{ij}=0$ otherwise.
\end{enumerate}
More details on the notations and the simulation scheme is presented in \citet{Lin2019}. In our simulation, the difference on the steps of data generation from that in \citet{Lin2019} is the addition of Step 6, which generates binary data. $\tc{T}_{lj}=1$ means gene $j$ is expressed in cell $l$, and $\tc{T}_{lj}=0$ otherwise. $\tc{A}_{ij}=1$ means the promoter region for feature $j$ is accessible in cell $i$, and $\tc{A}_{ij}=0$ otherwise.

\section{Details of feature selection for real data}
\label{ap:feature}
In the experiment, we implement variable selection for real single-cell genomic datasets before performing clustering, for the purpose of
speeding up computation and balances the number of variables among different data types. For real data 1, we apply clustering to the individual datasets \citep{Zama2018} and then select cluster-specific features \citep{Love2014, Zama2018}. Using the R toolkit Seurat \citep{Butler2018, Butler2019}, we select 100 most variable cluster-specific genes in each of scATAC-seq and scRNA-seq data. For real data 2, we also use Seurat to select the 100 most variable homologs from each of the mouse and human scRNA-seq data.

\end{appendices}
\end{document}